# The Bregman Variational Dual-Tree Framework

**Saeed Amizadeh**
Intelligent Systems Program
University of Pittsburgh
Pittsburgh, PA 15213

**Bo Thiesson**
Department of Computer Science
Aalborg University
Aalborg, Denmark

**Milos Hauskrecht**
Department of Computer Science
University of Pittsburgh
Pittsburgh, PA 15213

## Abstract

Graph-based methods provide a powerful tool set for many non-parametric frameworks in Machine Learning. In general, the memory and computational complexity of these methods is quadratic in the number of examples in the data which makes them quickly infeasible for moderate to large scale datasets. A significant effort to find more efficient solutions to the problem has been made in the literature. One of the state-of-the-art methods that has been recently introduced is the Variational Dual-Tree (VDT) framework. Despite some of its unique features, VDT is currently restricted only to Euclidean spaces where the Euclidean distance quantifies the similarity. In this paper, we extend the VDT framework beyond the Euclidean distance to more general Bregman divergences that include the Euclidean distance as a special case. By exploiting the properties of the general Bregman divergence, we show how the new framework can maintain all the pivotal features of the VDT framework and yet significantly improve its performance in non-Euclidean domains. We apply the proposed framework to different text categorization problems and demonstrate its benefits over the original VDT.

## 1 Introduction

Graph-based methods provide a powerful tool set for many non-parametric frameworks in Machine Learning (ML). The common assumption behind these methods is the datapoints can be represented as the nodes of a graph whose edges encode some notion of *similarity* between the datapoints. Graph-based methods have been applied to various applications in ML including clustering (Ng et al., 2001a; von Luxburg, 2007; Amizadeh et al., 2012b), semi-supervised learning (Zhu, 2005; Zhou et al., 2003), link-analysis (Ng et al., 2001c,b) and dimensionality reduction (Belkin and Niyogi, 2002; Zhang et al., 2012).

On the algorithmic side, many of these methods involve computing the *random walk* on the graph which is mathematically represented by a *(Markov) transition matrix* over the graph. In general, the computation of such matrix takes $O(N^2)$ time and memory, where $N$ is the problem size. As a result, for problems with large $N$, the direct computation of the transition matrix (or its variants) quickly becomes infeasible. This is indeed a big challenge for applying many graph-based frameworks specially with the advent of large-scale datasets in many fields in ML. To tackle this challenge, a significant effort has been made in the literature to develop the approximation techniques that somehow *reduce* the representation of the underlying graph. Based on the nature of approximation, these methods are generally categorized into node reduction (Kumar et al., 2009; Talwalkar et al., 2008; Amizadeh et al., 2011), edge reduction (von Luxburg, 2007; Jebara et al., 2009; Qiao et al., 2010), Fast Gauss Transform (Yang et al., 2003, 2005) and hierarchical (Amizadeh et al., 2012a; Lee et al., 2011) techniques.

Recently, Amizadeh et. al (Amizadeh et al., 2012a) have proposed a hierarchical approximation framework called the Variational Dual-Tree (VDT) framework for the same purpose. In particular, by combining the variational approximation with hierarchical clustering of data, VDT provides a fast and scalable method to *directly* approximate the transition matrix of the random walk. Furthermore, the hierarchical nature of VDT makes it possible to have approximations at different levels of granularity. In fact, by changing the level of approximation in VDT, one can adjust the trade-off between computational complexity and approximation accuracy. One major restriction with the VDT framework, however, is it only works for Euclidean spaces where similarity is expressed via the Euclidean distance. This can be problematic in many real applications where the Euclidean distance is not the best way to encode similarity. Unfortunately, the extension of VDT to a general distance metric is *not* straightforward, mainly because the key computational gain of VDT is achieved by relying on the functional form of the Euclidean distance.

In this paper, our goal is to extend the VDT framework to use a general class of divergences called *Bregman divergences* (Banerjee et al., 2005). We propose this new method as the *Bregman Variational Dual Tree (BVDT)* framework. Bregman divergences cover a diverse set of divergences and distances which can all be reformulated in a unified functional form. They have been used in many ML paradigms including clustering (Banerjee et al., 2005), matrix approximation (Dhillon and Sra, 2005; Banerjee et al., 2004), nearest neighbor retrieval (Cayton, 2008) and search (Zhang et al., 2009). From the applied side, some famous distances and divergences such as Euclidean distance, KL-Divergence and Logistic Loss are in fact the instances of Bregman divergences. By extending the VDT framework to Bregman divergences, one can use it with any instance of Bregman divergences depending on the application and therefore it becomes accessible to a large class of applications where similarity is expressed via non-Euclidean measures.

The crucial aspect of using Bregman divergences for the VDT framework is it does not cost us any extra order of computations; it still has the same computational and memory complexity order as the original VDT. In particular, we show that by exploiting the functional form of the general Bregman divergence, one can design a similar mechanism as in VDT to significantly cut unnecessary distance computations. This is a very important property because the motivation to develop variational dual-trees in the first place was to tackle large-scale problems.

One nice feature of the VDT framework is its probabilistic interpretation. In fact, the whole framework is derived from the data likelihood. We show that by using the natural correspondence between the Bregman divergences and the exponential families, we can reconstruct the same probabilistic interpretation for the BVDT framework which induces a more general modeling perspective for the problems we consider. The benefit of such probabilistic view is not restricted only to the theoretical aspects; as we show by a walk-through example, one can utilize the probabilistic view of our model to derive appropriate Bregman divergences for those domains where the choice of a good Bregman divergence is not apparent. In particular, in this paper, we use this construction procedure to derive a proper Bregman divergence for frequency data (specifically text data in our experiments). By applying the BVDT framework equipped with the derived divergence on various text datasets, we show the clear advantage of our framework over the original VDT framework in terms of the approximation accuracy, while preserving the same computational complexity. Finally, we show that the original VDT framework is in fact a special case of the BVDT framework.

## 2 Euclidean Variational Dual-Trees

To prepare the reader for the proposed Bregman divergence extension to the VDT framework, we will in this section briefly review the basic elements of VDT. Interested readers may refer to (Amizadeh et al., 2012a) for further details.

### 2.1 Motivation

Let $\mathcal{D} = \{x_1, x_2, \ldots, x_N\}$ be a set of i.i.d. datapoints in $\mathcal{X} \subseteq \mathbb{R}^d$. The similarity graph $\mathcal{G} = \langle \mathcal{D}, \mathcal{D} \times \mathcal{D}, \mathbb{W} \rangle$ is defined as a *complete* graph whose nodes are the elements of $\mathcal{D}$ and the matrix $\mathbb{W} = [w_{ij}]_{N \times N}$ represents the edge weights of the graph. A weight $w_{ij} = k(x_i, x_j; \sigma) = \exp(-\|x_i - x_j\|^2/2\sigma^2)$ is defined by the similarity between nodes $x_i$ and $x_j$. The closer $x_i$ and $x_j$ are in terms of the Euclidean distance $\|x_i - x_j\|$, the higher the similarity weight $w_{ij}$ will be. The bandwidth parameter $\sigma$ is a design parameter that scales the similarities in the graph. By abstracting the input space into a graph, the similarity graph essentially captures the geometry of the data.

Once the similarity graph is constructed, one can define a random walk on it. The transition probability distributions are in this case given by the transition matrix $\mathbb{P} = [p_{ij}]_{N \times N}$, where $p_{ij} = w_{ij} / \sum_{k \neq i} w_{ik}$ is the probability of jumping from node $x_i$ to node $x_j$. Note that the elements in each row of $\mathbb{P}$ sum up to 1 so that each row is a proper probability distribution. The transition matrix is the fundamental quantity used by many graph-based Machine Learning frameworks, as mentioned already in Section 1.

In terms of computational resources, it takes $O(N^2)$ CPU and memory to construct and maintain the transition matrix, which can be problematic when the number of datapoints, $N$, becomes large. This is, in fact, quite typical in many real-world datasets and therefore leaves us with a serious computational challenge in using many graph-based frameworks. To overcome this challenge, the key idea is to somehow *reduce* the representation of $\mathbb{P}$. The Variational Dual-Tree (VDT) framework provides a non-parametric methodology to approximate and represent

$\mathbb{P}$ in $O(N^{1.5} \log N)$. One big advantage of this framework to its counterparts is that it directly computes the transition matrix without computing the intermediate similarity matrix $\mathbb{W}$. Another advantage of VDT is that given a distance computation of $O(1)$ between any two datapoints, the overall computational complexity of VDT does not depend on the dimensionality $d$ of the input space.

## 2.2 Variational Dual-Tree Partitioning

The main idea behind the computational reduction by the VDT framework is to *partition* the transition matrix $\mathbb{P}$ into *blocks*, where each block ties the individual transition probabilities in that block into a single number (or parameter). The number of different elements in $\mathbb{P}$ is in this way reduced from $N^2$ to $|\mathcal{B}|$, the number of blocks. In the VDT framework, a cluster tree hierarchy of the data lets us define blocks of different sizes according to nodes at different granularity levels in the tree. In this way, the number of blocks $|\mathcal{B}|$ can be as small as $O(N)$ (Amizadeh et al., 2012a).

More specifically, let $\mathcal{T}$ be a binary tree that represents a hierarchical clustering of data. Given $\mathcal{T}$, a *valid* block partition $\mathcal{B}$ defines a mutually exclusive and exhaustive partition of $\mathbb{P}$ into blocks (or sub-matrices) $(A, B) \in \mathcal{B}$, where $A$ and $B$ are two *non-overlapping* subtrees in $\mathcal{T}$. That is, $A$ cannot be a subtree of $B$ or vice versa. A valid block partition $\mathcal{B}$ relates to a similarity graph as follows. If $A$ and $B$ are two non-overlapping datapoint clusters, then the block $(A, B) \in \mathcal{B}$ represents the transition probabilities from datapoints in node $A$ to datapoints in node $B$ with only one parameter, which is denoted by $q_{AB}$. That is, $\forall x_i \in A, x_j \in B, p_{ij} = q_{AB}$; we call this a *block constraint*. Valid block partitions are not unique; in fact, any further *refinement* of a valid partition results in a new valid partition with increased number of blocks. In the *coarsest* partition, each subtree in $\mathcal{T}$ is blocked with its sibling, resulting in the minimum number of blocks $|\mathcal{B}| = 2(N-1)$. The coarsest partition embodies the approximation of $\mathbb{P}$ at the coarsest level. On the other hand, in the *finest* partition, each leaf in $\mathcal{T}$ is blocked with all the other leaves resulting in the maximum number of blocks $|\mathcal{B}| = N(N-1)$; this partition exactly represents $\mathbb{P}$ with no approximation. The other valid partitions vary between these two extremes. In this setup, a finer partition has better approximation accuracy at the cost of increased computational complexity and vice versa. Figure 1 borrowed from (Amizadeh et al., 2012a) shows an example of a block partitioning with its corresponding cluster tree.

Given a block partitioning $\mathcal{B}$ of $\mathbb{P}$, one needs to compute the parameter set $\mathcal{Q} = \{q_{AB} \mid (A, B) \in \mathcal{B}\}$ as

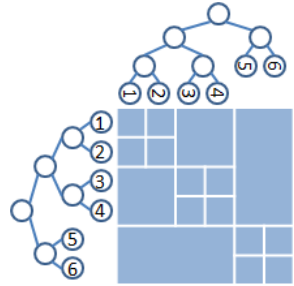

Figure 1: A block partition that, for example, enforces the block constraint $p_{13} = p_{14} = p_{23} = p_{24}$ for the block $(A, B) = (1-2, 3-4)$ (where $a-b$ denotes $a$ through $b$).

the final step to complete the approximation of $\mathbb{P}$. For this purpose, the VDT framework maximizes the variational lower bound on the log-likelihood of data with the set $\mathcal{Q}$ as the variational parameters. In this framework, the likelihood of data is modeled by the non-parametric kernel density estimate. In particular, a Gaussian kernel is placed on each datapoint in the input space such that each datapoint $x_i$ plays two roles: (I) a datapoint where we want to compute the likelihood at, and (II) the center of a Gaussian kernel. We denote datapoint $x_i$ as $m_i$ when it is regarded as a kernel center. Using this notation, the likelihood of dataset $\mathcal{D}$ is computed as:

$$p(\mathcal{D}) = \prod_i p(x_i) = \prod_i \sum_{j \neq i} p(m_j) p(x_i \mid m_j), \quad (1)$$

where $p(x_i|m_j)$ is the Gaussian density at $x_i$ centered at $m_j$; that is, $p(x_i|m_j) = \exp(-\|x_i - m_j\|^2/2\sigma^2)(2\pi\sigma^2)^{-d/2}$, and $p(m_j) = 1/(N-1)$ is the uniform mixture weight. Using Bayes rule, we observe that the posterior $p(m_j \mid x_i)$ is equal to the transition probability $p_{ij}$. The lower-bound on the log-likelihood is then computed as:

$$\begin{aligned} \log p(\mathcal{D}) &= \sum_i \log \sum_{j \neq i} \frac{q_{ij}}{q_{ij}} p(m_j) p(x_i \mid m_j) \\ &\geq \sum_i \sum_{j \neq i} q_{ij} \log \frac{p(m_j) p(x_i \mid m_j)}{q_{ij}} \\ &= \log p(\mathcal{D}) - \sum_i D_{KL}(q_{i\cdot} \| p_{i\cdot}) \triangleq \ell(\mathcal{D}), \quad (2) \end{aligned}$$

where $q_{ij}$'s are the variational parameters approximating $p_{ij}$'s and $D_{KL}(q_{i\cdot} \| p_{i\cdot})$ is the KL-divergence between two distributions $q_{i\cdot}$ and $p_{i\cdot}$. With only the sum-to-one constraints $\forall i : \sum_{j \neq i} q_{ij} = 1$, the optimization in Eq. (2) returns $q_{ij} = p_{ij}$; that is, there is no approximation! However, by adding the block constraints from the block partition $\mathcal{B}$, one can rewrite Eq. (2) in terms of the block parameters in $\mathcal{Q}$. Let us first

reformulate the sum-to-one constraints in accordance with the block partition as follows:

$$\sum_{(A,B)\in\mathcal{B}(x_i)} |B| \cdot q_{AB} = 1 \text{ for all } x_i \in \mathcal{D}. \quad (3)$$

where, $\mathcal{B}(x_i) \triangleq \{(A,B) \in \mathcal{B} \mid x_i \in A\}$. In this case,

$$\begin{aligned}\ell(\mathcal{D}) &= c - \frac{1}{2\sigma^2} \sum_{(A,B)\in\mathcal{B}} q_{AB} \cdot D_{AB} \\ &\quad - \sum_{(A,B)\in\mathcal{B}} |A||B| \cdot q_{AB} \log q_{AB},\end{aligned} \quad (4)$$

where

$$\begin{aligned}c &= -N \log\left((2\pi)^{d/2}\sigma^d(N-1)\right) \\ D_{AB} &= \sum_{x_i \in A} \sum_{m_j \in B} \|x_i - m_j\|^2.\end{aligned} \quad (5)$$

(Thiesson and Kim, 2012) has proposed an $O(|\mathcal{B}|)$-time algorithm to maximize Eq. (4) under the constraints in Eq. (3).

A very crucial element of the VDT framework is the way that Eq. (5) is computed; the direct computation of the double-sum for $D_{AB}$ of all blocks would send us back to an $O(N^2)$-time algorithm! Fortunately, this can be avoided thanks to the Euclidean distance, where $D_{AB}$ can be written as:

$$D_{AB} = |A|S_2(B) + |B|S_2(A) - 2S_1(A)^T S_1(B), \quad (6)$$

where, $S_1(A) = \sum_{x \in A} x$ and $S_2(A) = \sum_{x \in A} x^T x$ are the statistics of subtree $A$. These statistics can be incrementally computed and stored while the cluster tree is being built; an $O(N)$ computation. Using these statistics, $D_{AB}$ is computed in $O(1)$. The crux of the reformulation in Eq. (6) is the de-coupling of the *mutual interactions* between two clusters $A$ and $B$ so that the sum of mutual interactions can be computed using the sufficient statistics pre-calculated independently for each cluster.

Once the parameters $\mathcal{Q}$ are computed, we have the block approximation of $\mathbb{P}$ which we denote by $\mathbb{Q}$. (Amizadeh et al., 2012a) has proposed an $O(|\mathcal{B}|)$-time algorithm to compute the matrix-vector multiplication $\mathbb{Q}v$ for an arbitrary vector $v$. We will use this algorithm for label propagation in Eq. (29) in Section 4.2.

### 2.3 Anchor Tree Construction

So far, we assumed the cluster hierarchy tree $\mathcal{T}$ is given. In reality, however, one needs to efficiently build such a hierarchy from data as the first step of the VDT framework. (Amizadeh et al., 2012a) has used *the anchor construction* method (Moore, 2000) for this purpose. Compared to the classical $O(N^3)$-time agglomerative clustering algorithm, the construction time for this tree is $O(N^{1.5} \log N)$ for a relatively balanced data set (see Amizadeh et al. (2012a), Appendix). The construction starts with an *anchor growing phase* that for the $N$ data points gradually grows a set of $\sqrt{N}$ *anchors* $\mathcal{A}$. Each anchor $A \in \mathcal{A}$ has a *pivot* datapoint $A_p$ and maintains a list of individual *member* datapoints $A_M$ sorted by decreasing order of distance to the pivot $\|x_i - A_p\|$, $x_i \in A_M$. The distance to the first datapoint in the list therefore defines a *covering radius* $A_r$ for the anchor, where $\|x_i - A_p\| \leq A_r$ for all $x_i \in A_M$.

The anchor growing phase constructs a first anchor by choosing a random datapoint as pivot and assigning all datapoints as members of that anchor. A new anchor $A^{new}$ is now added to a current set of anchors in three steps until $\sqrt{N}$ anchors are found: first, its pivot element is chosen as the datapoint with largest distance to the pivot(s) of the current anchor(s):

$$A_p^{new} = \arg\max_{x_i \in A_M, A \in \mathcal{A}} \|x_i - A_p\|. \quad (7)$$

Second, the new anchor now iterates through the member elements of current anchors and *"steals"* the datapoints with $\|x_i - A_p^{new}\| < \|x_i - A_p\|$. Because the list of elements in an anchor is sorted with respect to $\|x_i - A_p\|$, a significant computational gain is achieved by not evaluating the remaining datapoints in the list once we discover that for the $i$-th datapoint in the list $\|x_i - A_p\| \leq d_{thr}$, where

$$d_{thr} = \|A_p^{new} - A_p\|/2, \quad (8)$$

This guarantees that the elements with index $j \geq i$ in the list are closer to their original anchor's pivot and cannot be stolen by the new anchor. When a new anchor is done stealing datapoints from older anchors, its list of elements is finally sorted.

Once the $\sqrt{N}$ anchors are created, the anchor tree construction now proceeds to *anchor agglomeration phase* that assigns anchors as leaf nodes and then iteratively merges two nodes that create a parent node with the *smallest* covering radius. This agglomerative bottom-up process continues until a hierarchical binary tree is constructed over the $\sqrt{N}$ initial anchors. With an Euclidean distance metric, the covering parent for two nodes $A$ and $B$ can be readily computed as the node $C$ with pivot and radius

$$\begin{aligned}C_p &= (|A| \cdot A_p + |B| \cdot B_p)/(|A| + |B|) & (9) \\ C_r &= (A_r + B_r + \|B_p - A_p\|)/2 & (10)\end{aligned}$$

Finally, recall that the leaves (i.e. the initial anchors) in this newly constructed tree contain $\sqrt{N}$ datapoints on average each. The whole construction algorithm is now recursively called for each anchor leaf to grow it into a subtree. The recursion ends when the leaves of the tree contain only one datapoint each.

## 3 Bregman Variational Dual-Trees

The Euclidean VDT framework has been shown to be a practical choice for large-scale applications. However, its inherent assumption that the underlying distance metric in the input space should be the Euclidean distance is somewhat restrictive. In many real-world problems, the Euclidean distance is simply not the best way to quantify the similarity between datapoints.

On the other hand, the VDT framework does not seem to depend on the choice of distance metric, which makes it very tempting to replace the Euclidean distance in the formulation of the VDT framework with a general distance metric. However, there is one problem: the de-coupling in Eq. (6) was achieved only because of the Euclidean distance special form which is not the case for a general distance metric. Unfortunately, we cannot compromise on this de-coupling simply because, without it, the overall complexity of the framework is back to $O(N^2)$. One solution is to use some approximation technique similar to Fast Gauss Transform techniques (Yang et al., 2005) to approximately de-couple a general distance metric. Although, this may work well for some special cases, in general, the computational burden of such approximation can be prohibitive; besides, we will have a new source of approximation error.

So, if we cannot extend VDT for general distance metric, is there any sub-class of metrics or divergences which we can safely use to extend the VDT framework? The answer is yes, the family of *Bregman divergences* is a qualified candidate. This family contains a diverse set of divergences which also include the Euclidean distance. By definition, the Bregman divergence has a de-coupled form which makes it perfect for our purpose. From the applied side, Bregman divergences cover some very practical divergences and metrics such as Euclidean distance, KL-Divergence and Logistic Loss that are widely used in many engineering and scientific applications. Furthermore, the natural correspondence of Bregman divergences with the exponential families provides a neat probabilistic interpretation for our framework.

### 3.1 Bregman Divergence and The Exponential Families

Before illustrating the Bregman Variational Dual-Tree (BVDT) framework, we briefly review the Bregman divergence, its important properties and its connection to the exponential families. Interested readers may refer to (Banerjee et al., 2005) for further details.

Let $\mathcal{X} \subseteq \mathbb{R}^d$ be a convex set in $\mathbb{R}^d$, $ri(\mathcal{X})$ denote the relative interior of $\mathcal{X}$, and $\phi : \mathcal{X} \mapsto \mathbb{R}$ be a strictly convex function differentiable on $ri(\mathcal{X})$, then the Bregman divergence $d_\phi : \mathcal{X} \times ri(\mathcal{X}) \mapsto [0, \infty)$ is defined as:

$$d_\phi(x, y) \triangleq \phi(x) - \phi(y) - (x - y)^T \nabla \phi(y) \quad (11)$$

where, $\nabla \phi(y)$ is the gradient of $\phi(\cdot)$ evaluated at $y$. For different choices of $\phi(\cdot)$, we will get different Bregman divergences. Table 1 lists some famous Bregman divergences along with their corresponding $\phi(\cdot)$ functions. It is important to note that the general Bregman divergence is *not* a distance metric: it is not symmetric, nor does it satisfy the triangular inequality. However, we have $\forall x \in \mathcal{X}, y \in ri(\mathcal{X}) : d_\phi(x, y) \geq 0, d_\phi(y, y) = 0$.

Let $\mathcal{S} = \{x_1, \ldots, x_n\} \subset \mathcal{X}$ and $X$ be a random variable that takes values from $\mathcal{S}$ with uniform distribution[1]; the *Bregman information* of the random variable $X$ for the Bregman divergence $d_\phi(\cdot, \cdot)$ is defined as:

$$I_\phi(X) \triangleq \min_{s \in ri(\mathcal{X})} \mathbb{E}\big[d_\phi(X, s)\big] = \min_{s \in ri(\mathcal{X})} \frac{1}{n} \sum_{i=1}^n d_\phi(x_i, s) \quad (12)$$

The optimal $s$ that minimizes Eq. (12) is called the *Bregman representative* of $X$ and is equal to:

$$\arg \min_{s \in ri(\mathcal{X})} \mathbb{E}\big[d_\phi(X, s)\big] = E[X] = \frac{1}{n} \sum_{i=1}^n x_i \triangleq \mu \quad (13)$$

That is, the Bregman representative of $X$ is always equal to the sample mean of $\mathcal{S}$ independent of the Bregman divergence $d_\phi(\cdot, \cdot)$.

The probability density function $p(z)$, defined on set $\mathcal{Z}$, belongs to an *exponential family* if there exists a mapping $g : \mathcal{Z} \mapsto \mathcal{X} \subseteq \mathbb{R}^d$ that can be used to re-parameterize $p(z)$ as:

$$p(z) = p(x; \theta) = \exp(\theta^T x - \psi(\theta)) p_0(x) \quad (14)$$

where $x = g(z)$ is the *natural statistics* vector, $\theta$ is the *natural parameter* vector and $\psi(\theta)$ is the *log-partition function*. Eq. (14) is called the *canonical form* of $p$. If $\theta$ takes values from parameter space $\Theta$, Eq. (14) defines the family $\mathcal{F}_\psi = \{p(x; \theta) \mid \theta \in \Theta\}$ parameterized by $\theta$. If $\Theta$ is an open set and we have that $\nexists c \in \mathbb{R}^d$ s.t. $c^T g(z) = 0, \forall z \in \mathcal{Z}$, then family $\mathcal{F}_\psi$ is called a *regular exponential family*. (Banerjee et al., 2005) has shown that any probability density function $p(x; \theta)$ of a regular exponential family with the canonical form of Eq. (14) can be uniquely expressed as:

$$p(x; \theta) = \exp(-d_\phi(x, \mu)) \exp(\phi(x)) p_0(x) \quad (15)$$

where $\phi(\cdot)$ is the *conjugate function* of the log-partition function $\psi(\cdot)$, $d_\phi(\cdot, \cdot)$ is the Bregman divergence defined w.r.t. function $\phi(\cdot)$, and $\mu$ is the *mean parameter*. The mean parameter vector $\mu$ and the natural

---
[1] The results hold for any distribution on $\mathcal{S}$.

parameter vector $\theta$ are connected through:

$$\mu = \nabla\psi(\theta), \quad \theta = \nabla\phi(\mu) \quad (16)$$

Moreover, (Banerjee et al., 2005) (Theorem 6) has shown there is a bijection between the regular exponential families and the regular Bregman divergences. The last column in Table 1 shows the corresponding exponential family of each Bregman divergence. Using Eq. (13), one can also show that, given the finite sample $\mathcal{S} = \{x_1, \ldots, x_n\}$, the maximum-likelihood estimate of the mean parameter $\hat{\mu}$ for any regular exponential family $\mathcal{F}_\psi$ is always equal to the sample mean of $\mathcal{S}$ regardless of $\mathcal{F}_\psi$.

## 3.2 Bregman Variational Approximation

Having described the basic concepts of Bregman divergences, we are now ready to nail down the BVDT framework. Let $\mathcal{S} = \{z_1, z_2, \ldots, z_N\} \subset \mathcal{Z}$ be a finite sample from the convex set $\mathcal{Z}$ which is not necessarily an Euclidean space. We are interested to approximate the transition matrix $\mathbb{P}$ on the similarity graph of $\mathcal{S}$ where we know the Euclidean distance is not necessarily the best way to encode similarity. To do so, we assume $\mathcal{S}$ is sampled according to an unknown mixture density model $p^*(z)$ with $K$ components from the regular exponential family $\mathcal{F}_\psi$. That is, there exists the mapping $g : \mathcal{Z} \mapsto \mathcal{X} \subseteq \mathbb{R}^d$ such that $p^*(z)$ can be re-parametrized in the canonical form as $p^*(x) = \sum_{i=1}^{K} p(\theta_i) \exp(\theta_i^T x - \psi(\theta_i)) p_0(x)$.[2]

Furthermore, let $\mathcal{D} = g(\mathcal{S}) = \{x_1, x_2, \ldots, x_N\} \subset \mathcal{X}$ be the natural statistics of sample $\mathcal{S}$ s.t. $x_i = g(z_i)$. Then we model the likelihood of $\mathcal{D}$ using the kernel density estimation:

$$p(\mathcal{D}) = \prod_{i=1}^{N} \sum_{j \neq i} p(m_j) p(x_i \mid m_j) \quad (17)$$

$$= \prod_{i=1}^{N} \sum_{j \neq i} p(m_j) \exp(-d_\phi(x_i, m_j)) \exp(\phi(x_i)) p_0(x_i)$$

Where, we assume the kernel component $p(x_i \mid m_j)$ belongs to the regular exponential family $\mathcal{F}_\psi$ such that it can be uniquely re-parameterized using Eq. (15). Given a block partitioning $\mathcal{B}$ on $\mathbb{P}$, we follow the similar steps in Eq. (2)-(5) to derive the block-partitioned variational lower-bound on $p(\mathcal{D})$:

$$\ell(\mathcal{D}) = c - \sum_{(A,B)\in\mathcal{B}} q_{AB} \cdot D_{AB}$$

$$- \sum_{(A,B)\in\mathcal{B}} |A||B| \cdot q_{AB} \log q_{AB}, \quad (18)$$

---
[2]For some exponential families such as Gaussian and Multinomial, the mapping $g(\cdot)$ is identity.

where

$$c = -N\log(N-1) + \sum_{i=1}^{N}\bigl(\phi(x_i) + \log p_0(x_i)\bigr)$$

$$D_{AB} = \sum_{x_i \in A} \sum_{m_j \in B} d_\phi(x_i, m_j). \quad (19)$$

Now we can maximize $\ell(\mathcal{D})$ subject to the constraints in Eq. (3) to find the approximation $\mathbb{Q}$ of $\mathbb{P}$ using the same $O(|\mathcal{B}|)$-time algorithm in the VDT framework.

The crucial aspect of the BVDT framework is $D_{AB}$ in Eq. (19) is de-coupled into statistics of the subtrees $A$ and $B$ using the definition of the Bregman divergence:

$$D_{AB} = |B|S_1(A) + |A|\bigl(S_2(B) - S_1(B)\bigr) - S_3(A)^T S_4(B) \quad (20)$$

where,

$$S_1(A) = \sum_{x \in A} \phi(x), \quad S_2(A) = \sum_{x \in A} x^T \nabla\phi(x)$$

$$S_3(A) = \sum_{x \in A} x, \quad S_4(A) = \sum_{x \in A} \nabla\phi(x) \quad (21)$$

are the statistics of subtree $A$. These statistics can be incrementally computed and stored while the cluster tree is being built (in overall $O(N)$ time) such that at the optimization time, $D_{AB}$ is computed in $O(1)$.

Finally, by setting $\phi(x) = \|x\|^2/2\sigma^2$ and doing the algebra, the BVDT framework reduces to the Euclidean VDT framework; that is, the Euclidean VDT framework is a special case of the BVDT framework.

## 3.3 Bregman Anchor Trees

Recall that the approximation in the Euclidean VDT framework is based on the cluster hierarchy $\mathcal{T}$ of the data which is built using the anchor tree method with the Euclidean distance. For the BVDT framework, we can no longer use this algorithm because the Euclidean distance no longer reflects the similarity in the input space. For this reason, one needs to develop an anchor tree construction algorithm for general Bregman divergences. This generalization is not straightforward as a general Bregman divergence is neither symmetric nor does it hold the triangle inequality. In particular, we need to address two major challenges.

First, due to the asymmetry of a general Bregman divergence, the merging criterion in the anchor agglomeration phase is no longer meaningful. For this purpose, we have used the criterion suggested in the recent work by (Telgarsky and Dasgupta, 2012). In particular, at each agglomeration step, anchors $A$ and $B$ with the minimum *merging cost* are picked to merge into

| Name | $\mathcal{X}$ | $\phi(\mathbf{x})$ | $d_\phi(\mathbf{x}, \mathbf{y})$ | **Exponential Family** |
|---|---|---|---|---|
| **Logistic Loss** | $(0,1)$ | $x \log x$ | $x \log\left(\frac{x}{y}\right) + (1-x) \log\left(\frac{1-x}{1-y}\right)$ | 1D Bernoulli |
| **Itakura-Saito Dist.** | $\mathbb{R}_{++}$ | $-\log x - 1$ | $\frac{x}{y} - \log\left(\frac{x}{y}\right) - 1$ | 1D Exponential |
| **Relative Entropy** | $\mathbb{Z}_+$ | $x \log x - x$ | $x \log\left(\frac{x}{y}\right) - x + y$ | 1D Poisson |
| **Euclidean Dist.** | $\mathbb{R}^d$ | $\|x\|^2/2\sigma^2$ | $\|x-y\|^2/2\sigma^2$ | Spherical Gaussian |
| **Mahalonobis Dist.** | $\mathbb{R}^d$ | $x^T \Sigma^{-1} x$ | $(x-y)^T \Sigma^{-1} (x-y)$ | Multivariate Gaussian |
| **KL-Divergence** | $d$-simplex | $\sum_{j=1}^d x(j) \log x(j)$ | $\sum_{j=1}^d x(j) \log\left(\frac{x(j)}{y(j)}\right)$ | - |
| - | int. $d$-simplex | $\sum_{j=1}^d x(j) \log\left(\frac{x(j)}{L}\right)$ | $\sum_{j=1}^d x(j) \log\left(\frac{x(j)}{y(j)}\right)$ | Multinomial |

Table 1: Famous Bregman divergences along with their corresponding $\phi(\cdot)$ function, its domain and the corresponding exponential family distribution

the parent anchor $C$. The cost for merging a pair of anchors $A$ and $B$ is defined as:

$$\Delta(A,B) = |A| \cdot d_\phi(A_p, C_p) + |B| \cdot d_\phi(B_p, C_p) \quad (22)$$

where $|A|$ is the number of elements in the anchor $A$, and $C_p$ is the parent anchor's pivot which is given by Eq. (9). (Telgarsky and Dasgupta, 2012) has shown that the merging cost in Eq. (22) can be interpreted as the difference of cluster unnormalized Bregman informations before and after merging.

Second, as shown before, using the halfway Euclidean distance as the stealing threshold (Eq. (8)) in the anchor construction phase significantly cuts the unnecessary computations. This threshold, however, is meaningless for a general Bregman divergence simply because a general Bregman divergence is not a metric. Therefore, we need to develop an equivalent threshold for Bregman divergences to achieve a similar computational gain in constructing Bregman anchor trees. The following proposition addresses this problem:

**Preposition 1.** *Let $A^{curr}$ and $A^{new}$ denote the current and the newly created anchors, respectively, where $A^{new}$ is stealing datapoints from $A^{curr}$. Define*

$$d_{thr} = \frac{1}{2} \min_{y \in \mathcal{X}} \left[ d_\phi(y, A_p^{curr}) + d_\phi(y, A_p^{new}) \right] \quad (23)$$

*Then for all $x \in A^{curr}$ such that $d_\phi(x, A_p^{curr}) \le d_{thr}$, we will have $d_\phi(x, A_p^{curr}) \le d_\phi(x, A_p^{new})$; that is, $x$ cannot be stolen from $A_p^{curr}$ by $A_p^{new}$. Furthermore, the minimizer of Eq. (23) is equal to:*

$$y^* = \nabla \phi^{-1} \left[ \frac{1}{2} \left( \nabla \phi(A_p^{curr}) + \nabla \phi(A_p^{new}) \right) \right] \quad (24)$$

The proof is easily derived by contradiction. Note that for the special case of Euclidean distance where $\phi(x) = \|x\|^2/2\sigma^2$, Eq. (23) reduces to Eq. (8).

## 4 Experiments

In order to apply the BVDT framework to a real problem, all one needs to know are the appropriate Bregman divergence and its corresponding $\phi(\cdot)$ function in the input space; knowing the corresponding exponential family that has generated the data is *not* necessary. However, in many problems, the choice of the appropriate Bregman divergence is not clear; instead, we know that the underlying data generation process belongs to *or* can be accurately modeled with exponential families. In such cases, one can systematically derive the appropriate Bregman divergence from the data distribution. In this section, we use this procedure to derive an appropriate Bregman divergence for the frequency data.

### 4.1 Modeling Frequency Data

Given a set of $d$ events $\{e_j\}_{j=1}^d$, the frequency dataset $\mathcal{D}$ consists of $N$ feature vectors where the $j$-th element of the $i$-th vector, $x_i(j)$, represents the number of times that event $e_j$ happens in the $i$-th case for all $j \in \{1, \ldots, d\}$. The length of each vector is defined as the total number of events happened in that case, i.e. $L_i = \sum_{j=1}^d x_i(j)$. For example, in text analysis, events can represent all the terms that appeared in a text corpus, while the data cases represent the documents. The frequency dataset in this case is the famous bag-of-words. We adopt the term-document analogy for the rest of this section.

If the lengths of all documents were equal, one could model the document generation process with a mixture of multinomials, where each mixture component had different term generation probabilities while sharing the same length parameter (Banerjee et al., 2005). However, having the same length is not the case for many real datasets. To address this issue, we propose a mixture model for data generation whose $k$-th component ($k \in [1..K]$) is modeled by the following generative model:

$$p_k(x; \alpha_k, \lambda_k) = p(x \mid L; \boldsymbol{\alpha}_k) p(L; \lambda_k) \quad (25)$$

where, the length of a document, $L$, has a poisson distribution $p(L; \lambda_k)$ with mean length $\lambda_k$ and given $L$, document $x$ has a multinomial distribution $p(x \mid$

$L; \boldsymbol{\alpha}_k)$ with the term probabilities $\boldsymbol{\alpha}_k = [\alpha_k(j)]_{j=1}^d$. By doing the algebra, $p_k(x; \alpha_k, \lambda_k)$ can be written in the form of Eq. (14), where we have:

$$\theta_k = [\theta_k(j)]_{j=1}^d = \left[\log(\lambda_k \alpha_k(j))\right]_{j=1}^d,$$

$$\psi(\theta_k) = \sum_{j=1}^d \exp(\theta_k(j)), p_0(x) = \left(\prod_{j=1}^d x(j)!\right)^{-1} \quad (26)$$

That is, our generative model also belongs to an exponential family. By deriving the conjugate function of $\psi(\cdot)$, we get the $\phi(\cdot)$ function and its corresponding Bregman divergence as:

$$\phi(x) = \sum_{j=1}^d x(j) \log x(j) - \sum_{j=1}^d x(j) \quad (27)$$

$$d_\phi(x, y) = \sum_{j=1}^d \left[ x(j) \log\left(\frac{x(j)}{y(j)}\right) - x(j) + y(j) \right] \quad (28)$$

The divergence in Eq. (28) is called the *Generalized I-Divergence* (GID) which is a generalization of KL-Divergence (Dhillon and Sra, 2005). As a result, to work with frequency data (in particular text data in our experiments), we customize the BVDT framework to use GID as its Bregman divergence. It should be noted that GID is not the only non-Euclidean similarity measure between documents, other techniques such as co-citation (Šingliar and Hauskrecht, 2006) has been used before.

### 4.2 Experimental Setup

To evaluate the BVDT framework, we use it for a semi-supervised learning (SSL) task on various text datasets. (Amizadeh et al., 2012a) has experimentally shown the VDT framework can be well scaled to large-scale problems. Our framework has exactly the same order of complexity as VDT. Therefore, due to the space limit, we focus our evaluation only on the *accuracy* of SSL for text data. In particular, we show that while enjoying the same computational speed-up as the Euclidean VDT framework, the BVDT framework equipped with GID significantly improves the quality of learning over the VDT framework for text data.

For the SSL task, we want to *propagate* the labels from a small set of labeled examples to the rest of unlabeled examples over the similarity graph built over the examples. To do so, we use the following iterative propagation scheme (Zhou et al., 2003):

$$y^{(t+1)} \leftarrow \alpha \mathbb{M} y^{(t)} + (1-\alpha) y^0 \quad (29)$$

where $y^{(t)} \in \mathbb{R}^{N \times 1}$ is the vector of labels at time $t$, $\mathbb{M} \in \mathbb{R}^{N \times N}$ is the transition matrix $\mathbb{P}$ (or its approximation $\mathbb{Q}$), $y^0$ is the vector of initial partial labeling and $\alpha \in [0, 1]$ is the mixing coefficient. For all of our experiments, we iterate through this process 300 times with $\alpha = 0.01$. Upon completion of propagation, we compute the classification accuracy over the unlabeled data w.r.t. the held-out true labels. Note that this process is for binary classification problems; for problems with multiple classes, we perform one-vs-all scheme for each class and take the maximum.

We have compared four methods for computing matrix $\mathbb{M}$: (a) Euclidean VDT, (b) BVDT equipped with GID, (c) Exact method with Euclidean distance and (d) Exact method with GID. For Exact methods, the transition matrix $\mathbb{P}$ is exactly computed using the direct method. Due to memory limitations on our single machine, we could apply Exact methods only for the smaller datasets. For variational methods, we have used the coarsest level of approximation for the transition matrix; that is, $\mathbb{P}$ is approximated with only $2(N-1)$ number of parameters (or blocks).

We have applied these methods on five text datasets represented as bag-of-words used before in (Greene and Cunningham, 2006; Deng et al., 2011; Maas et al., 2011; Lang, 1995). Table 2 illustrates the details.

| Dataset | N | d | C |
|---|---|---|---|
| **BBC-Sport News Articles** | 737 | 4613 | 5 |
| **BBC News Articles** | 2225 | 9636 | 5 |
| **20 Newsgroup** | 11269 | 61188 | 20 |
| **NSF Research Abstracts** | 16405 | 18674 | 10 |
| **Large Movie Reviews** | 50000 | 89527 | 2 |

Table 2: Datasets used: $N$ = number of documents, $d$ = number of terms, $C$ = number of classes

### 4.3 Results

Figures 2(A-E) show the classification accuracy vs. the percentage of labeled data for the datasets. The plots show the average of 5 trials with 95% confidence intervals. Although, none of the actual datasets is generated by the generative model proposed in Eq. (25), the BVDT with GID method consistently and significantly outperforms the Euclidean VDT. In particular, these results show that (a) the Generalized I-Divergence derived from the proposed generative model for text data captures the document similarity much better than the Euclidean distance does, and (b) the BVDT framework provides a straightforward mechanism to extend the variational dual-tree method beyond the Euclidean distance to use Bregman divergences such as GID.

We have also applied the Exact methods to the two smallest datasets. Not surprisingly, the Exact method with GID has the best performance compared to other methods. However, the Exact method with a wrong distance metric (the Euclidean distance in this case) can do even worse than the VDT method with Euclidean distance, as observed for the second BBC

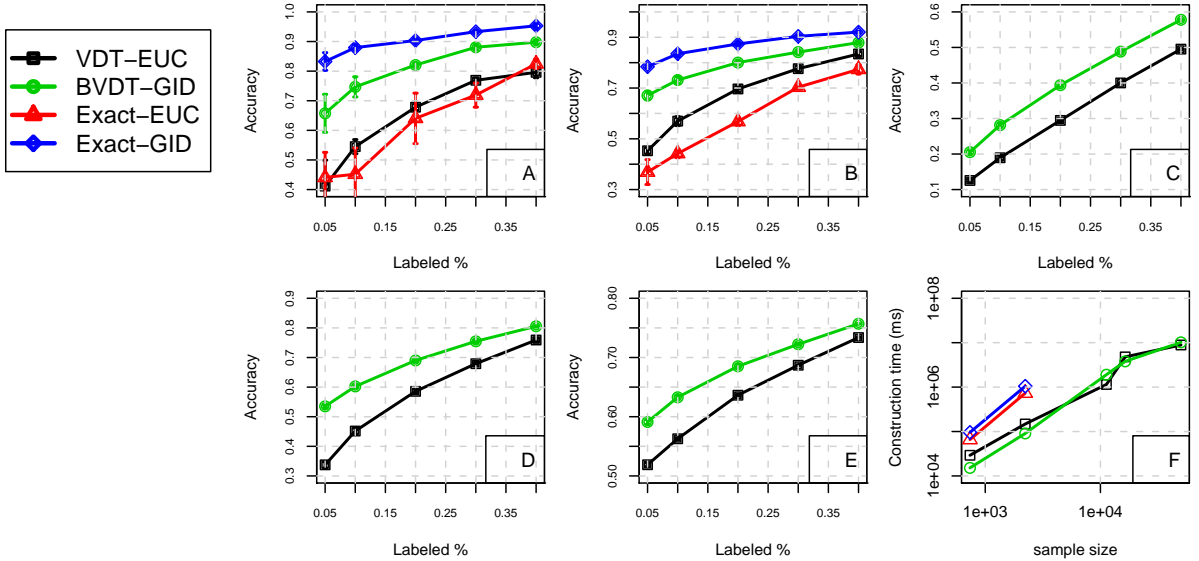

Figure 2: The accuracy curves vs. labeled data % for (A) BBC Sport News (B) BBC News (C) 20 Newsgroup (D) NSF Research Abstracts (E) Large Movie Reviews. (F) The computational complexity of four methods vs. the dataset size

dataset. We conjecture the existence of block regularization in the VDT framework compensates for the improper distance to some degree in this case.

Finally, we note the computational complexity of the aforementioned methods vs. the dataset size (i.e. the number of documents) shown in Figure 2(F). Both X an Y axes in this plot are in log-scale. As the plot shows while Euclidean VDT and BVDT have the same order complexity, they both are orders of magnitude faster than the Exact methods. In other words, while significantly improving on learning accuracy, BVDT still enjoys the same computational benefits as VDT.

## 5   Conclusions

In this paper, we proposed the Bregman Variational Dual-Tree framework which is the generalization of the recently developed Euclidean Variational Dual-Tree method to Bregman divergences. The key advantage of the BVDT framework is it covers a large class of distances and divergences and therefore makes the variational dual-trees accessible to many non-Euclidean large-scale datasets. The crucial aspect of our generalization to Bregman divergences is, unlike generalizing VDT to a general distance metric, it comes with no extra computational cost; that is, its computational order can still be kept the same as that of the VDT framework. This is very important to the development of whole framework since the variational dual-trees were originally developed to tackle large-scale problems. To achieve this, we utilized the functional form of the general Bregman divergence to design a bottom-up mechanism to cut unnecessary distance computations similar to that of the Euclidean VDT framework.

Furthermore, by exploiting the connection between the Bregman divergences and the exponential families, we provided a probabilistic view of our model. By a walk-through example, we showed that this probabilistic view can be used to derive the appropriate Bregman divergence for those domains where the best choice of distance in not apparent at the first glance. This example provides us with a powerful construction procedure to develop the appropriate Bregman divergence for a given problem. Specifically, we used this procedure to derive the Generalized I-Divergence for the frequency data. We showed that by incorporating GID in the BVDT framework, our model significantly improved the accuracy of learning for semi-supervised learning on various text datasets, while maintaining the same order of complexity as the original VDT framework. It should be emphasized that although we used the BVDT framework with one type of Bregman divergence, the proposed model is general and can be customized with any Bregman divergence.

## Acknowledgements

This research work was supported by grants 1R01LM010019-01A1 and 1R01GM088224-01 from the NIH. The content of this paper is solely the responsibility of the authors and does not necessarily represent the official views of the NIH.